\documentclass[letterpaper, 10 pt, conference]{ieeeconf} 
\usepackage{gensymb}
\usepackage{hyperref}
\usepackage{tabularx}
\usepackage{tabu}
\usepackage{graphicx}
\usepackage{caption}
\usepackage{subfigure}
\usepackage{color}
\usepackage{multirow}
\usepackage{amsmath}
\usepackage{array}
\usepackage{pifont}
\usepackage[noadjust]{cite}
\usepackage{amssymb}
    
\usepackage{graphicx,calc}
\newlength\myheight
\newlength\mydepth
\settototalheight\myheight{Xygp}
\newcommand*\inlinegraphics[1]{%
  \settototalheight\myheight{Xygp}%
  \settodepth\mydepth{Xygp}%
  \raisebox{-\mydepth}{\includegraphics[height=\myheight]{#1}}%
}
\usepackage{amsfonts}

\usepackage{soul,color}

\IEEEoverridecommandlockouts                             
\title{\LARGE \bf
NEMO: Future Object Localization Using Noisy Ego Priors
}
\author{Srikanth Malla \and Isht Dwivedi \and Behzad Dariush \and Chiho Choi
\thanks{The authors are with Honda Research Institute USA, 70 Rio Robles, San Jose, CA 95134, USA.
{\tt\small \{smalla,idwivedi, bdariush,cchoi\}@honda-ri.com}}
}
\newcommand{\norm}[1]{\left\lVert#1\right\rVert}
\begin{document}
\maketitle
\thispagestyle{empty}
\pagestyle{empty}

\begin{abstract}
Predicting the future trajectory of agents from visual observations is an important problem for realization of safe and effective navigation of autonomous systems in dynamic environments.   This paper focuses on two important aspects of future trajectory forecast which are particularly relevant for mobile platforms:  1) modeling uncertainty of the predictions, particularly from egocentric views, where uncertainty in the interactive reactions and behaviors of other agents must consider the uncertainty in the ego-motion, and 2) modeling multi-modality nature of the problem, which are particularly prevalent at junctions in urban traffic scenes.  To address these problems in a unified approach, we propose NEMO (Noisy Ego MOtion priors for future object localization) for future forecast of agents in the egocentric view. In the proposed approach, a predictive distribution of future forecast is jointly modeled with the uncertainty of predictions. For this, we divide the problem into two tasks: \textit{future ego-motion prediction} and \textit{future object localization}. We first model the  multi-modal distribution of future ego-motion with uncertainty estimates. The resulting distribution of ego-behavior is used to sample multiple modes of future ego-motion. Then, each modality is used as a prior to understand the interactions between the ego-vehicle and target agent. We predict the multi-modal future locations of the target from individual modes of the ego-vehicle while modeling the uncertainty of the target's behavior. To this end, we extensively evaluate the proposed framework using the publicly available benchmark dataset (HEV-I) supplemented with odometry data from an Inertial Measurement Unit (IMU).

\end{abstract}


\section{Introduction}
Predicting the future motion of agents in dynamic environments is a critical component for deployment of navigation and control strategies for autonomous and semi-autonomous systems in next generation mobility systems~\cite{introduction1,introduction2,introduction3}.  In particular, for safety critical applications that involve collision mitigation, forecasting future trajectory of agents in the scene must effectively model interactions between agents and account for the ego-motion, scene context, and environment constraints.   Although some approaches~\cite{yagi2018future,brian_icra,nikhil2018convolutional} have proposed a deterministic solution based on current and past history of the agents' motion, future forecast is inherently multi-modal, particularly  where multiple paths are plausible.


Other approaches~\cite{social_lstm,lee2017desire,social_gan,social_attention,drogon} have focused on modeling a distribution of all possible paths to tackle the multi-modality of future forecast. However, their predictive distribution is either (i) naively learned in a data driven manner with no consideration for the uncertainty; or (ii) simply generated to sample different types of motion using deep generative models. To provide a more compelling and rigorous solution to the multi-modality of the problem, a distribution of the agents' behavior should be jointly learned with uncertainty estimates, including uncertainty in the data (\textit{i.e.}, aleatoric), as well as uncertainty in  the prediction model (\textit{i.e.}, epistemic). 
In this way, the predicted modalities can quantify the level of uncertainty/noise, thereby increasing the confidence in the accuracy of the predicted future motions generated from the predictive distribution within the modality.

Research efforts in \textit{single-modal} future forecast~\cite{choi2019looking,cvpr_uncertainty} have shown that uncertainty embedding improves the overall performance for predicting the agents' future motion. However, \cite{choi2019looking} restricts their uncertainty to be epistemic, and overlooks noise inherent in the dataset, which makes it infeasible to recover from a small number of observations. Also, their problem setting (\textit{i.e.}, aerial RGB imagery as input) does not consider egomotion from a mobile platorform, making it difficult to deploy to autonomous driving (AD) and advanced driving assistance systems (ADAS). In~\cite{cvpr_uncertainty}, both aleatoric and epistemic uncertainty are considered from the ego-car perspective, where ego-motion as a prior affects the future motion of other agents. However, the uncertainty of ego-motion prediction is not taken into account, which is most critical to accurately forecast the interactive behavior of other agents.  

To address the limitations of existing approaches, we propose a \textit{multi-modal} future forecast framework, NEMO, which aims to (i) model both aleatoric and epistemic uncertainty of ego-vehicle as well as other agents; (ii) 
condition future object localization on multiple modes of ego-motion priors, which results in different types of target agents' behavior; 
and (iii) apply such a framework to immediate applications of autonomous driving (AD) and ADAS with readily  equipped  front-facing RGB camera. 

\begin{figure}[t]
    \includegraphics[width=0.49\textwidth]{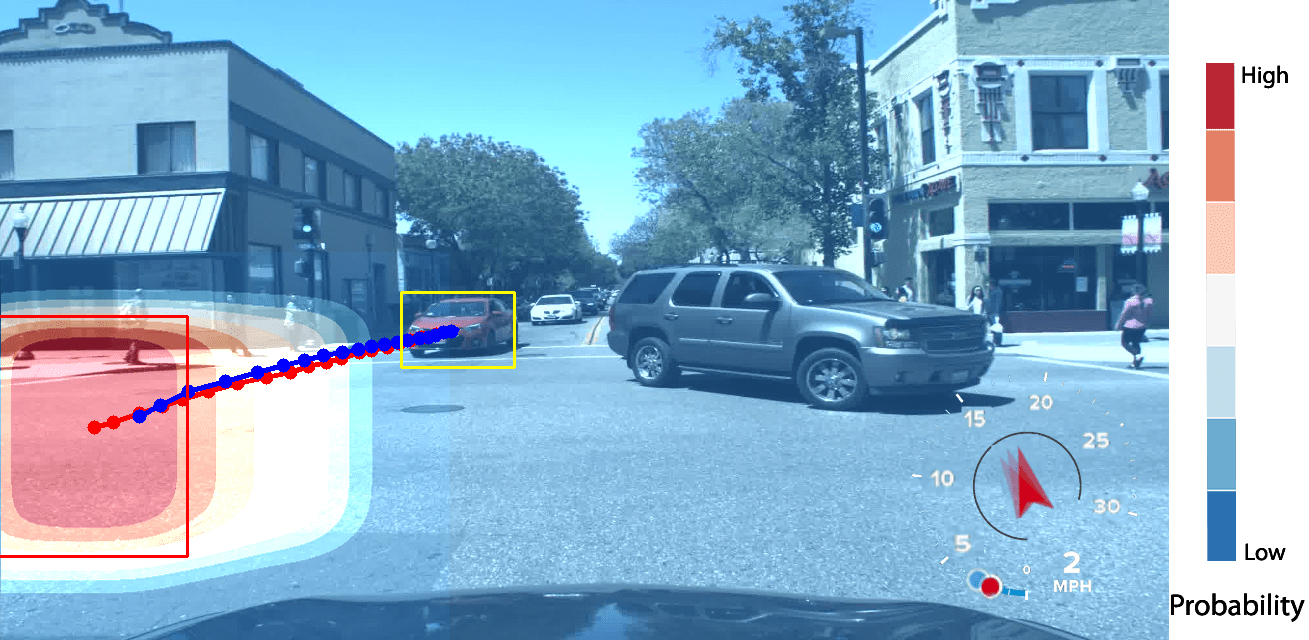}
    \caption{
    Given the uncertainty of future ego-motion over the velocity and yaw rate as a prior, the future trajectory and bounding box of the target agent is sampled from its uncertainty distribution. 
    }
    \label{fig:framework}
    \vspace{-0.51cm}
\end{figure}

Fig.~\ref{fig:framework} is a visual illustration of the proposed framework.  Our approach (NEMO) first models both the aleatoric and epistemic uncertainty of future ego-behavior using the past motion history of the ego-vehicle. Then, the multiple modes of future ego-motion are sampled from the probability distribution with uncertainty estimates. Each modality is provided to the future object localization stream as a prior to assess interactive responses of the target agent with respect to the different types of future ego-motion. We further consider the uncertainty of target agent's future motion and its multi-modality. An overview of the proposed approach is presented in Fig.~\ref{fig:block_diagram}. 
In this process, NEMO generates multi-modal future motions of the target over the uncertainty of future ego-motion, which reflects actual egocentric interactions observed in real traffic scenes. For more accurate ego-motion prediction,  IMU data synchronized with the video streams in  HEV-I~\cite{brian_icra} is released, which can extend the utility of HEV-I beyond future object localization problem to include visual odometry estimate~\cite{patil2019h3d} and other 2D image-based decision making and motion planning tasks~\cite{imagecontrol1,imagecontrol2}. The updated IMU sensor data will be made available at \url{https://usa.honda-ri.com/hevi}

\section{Related Work}
\subsection{Uncertainty Modeling}
Denker et al. \cite{denker1991transforming} and MacKay et al. \cite{mackay1992practical} studied the uncertainty of the model parameters using Bayesian neural networks (BNNs). Recently, Gal et al. \cite{variational_inference,gal2016dropout} have shown that Bayesian inference can be approximated with a traditional network architecture. They model epistemic uncertainty by sampling from the posterior distribution of the learned model using dropout during inference, which is equivalent to approximated Bayesian inference. In addition, Kendall et al. \cite{kendall_nips_uncertainty} shows that aleatoric uncertainty can be captured using negative log-likelihood loss by outputting the extra parameters for variance from the network output. It enables the network to learn the noise parameters that originate from the noise inherent in the dataset. Following the success of uncertainty modeling in single-modal forecast \cite{choi2019looking,cvpr_uncertainty}, we embed the uncertainty of future prediction into our multi-modal pipeline.

\subsection{Egocentric Vision}
Videos captured from the egocentric perspective are easy-available and contain the natural interactions of the ego-agent with the surrounding environment and other agents. Egocentric videos have been widely used in various tasks such as object detection \cite{detection1,detection2}, person re-identification \cite{reid1,reid2,reid3}, video summarization \cite{vidsummarization}, gaze prediction \cite{gazepredict}, and action recognition \cite{action1,action2,action3,action4,action5}. Recent works have looked into ego-action estimation using ego-view. Park et al. \cite{soo2016egocentric} studied future ego-location estimation using egocentric view. Su et al. \cite{su2017predicting} predict future actions for basketball players captured from synchronized multiple views using siamese networks.

The studies in~\cite{yagi2018future,brian_icra,cvpr_uncertainty} are directly related to future object localization in  first-person view. Yagi et al.~\cite{yagi2018future} uses
human poses as a prior to forecast the future motion of humans, but their model is not applicable to vehicles in traffic scenes. The work by \cite{brian_icra,cvpr_uncertainty} consider driving scenarios, but focus on single-modal localization of road agents.  In particular, Yao et al. \cite{brian_icra} uses object appearance and ground-truth future ego-motion for future localization, but does not consider prediction uncertainty. 
Battacharya et al. \cite{cvpr_uncertainty} predict the future ego-motion and use the prediction as prior to localize other agents with uncertainty estimates. However, their approach overlooks the uncertainly in the future ego-motion, which is critical in determining the interactive reactions of other agents and their future behaviors toward the ego-vehicle. In contrast, we address the uncertainty of future ego-motion prediction and introduce noisy ego-priors for multi-modal future object localization. 

\subsection{Future Trajectory Forecast}
The problem of future trajectory forecast from top-down views has been widely studied. Social-LSTM \cite{social_lstm} introduces a social pooling module for interaction encoding, and Social-GAN \cite{social_gan} efficiently improves its performance by replacing the pooling with a multi-layer perceptron. Social-Attention \cite{social_attention} introduces a soft attention mechanism to find more useful interactions. Gated-RN \cite{choi2019looking} observes spatio-temporal interactions using images and infers relational behavior between agents. Their relational inference is adopted into DROGON \cite{drogon} that uses intention as a prior for trajectory prediction, focusing on causation between intention and the future motion. SSP~\cite{dwivedi2020ssp} predicts all agents trajectories in single-shot using composite fields.

However, these methods are not seamlessly applicable to egocentric videos captured from mobile vehicle platforms for the following reasons: (i) unlike the top-down view from a stationary camera, the distance between objects should be jointly assumed from the location and scale in frontal view images; (ii) the interactions between agents are relative to the ego-motion, but their models do not explicitly account for the ego-motion uncertainty; and (iii) their consideration of the uncertainty does not exist or is minimal, which does not provide a comprehensive solution to multi-modal predictions. To address these limitations, we present NEMO for future trajectory forecast from an egocentric view.

\section{Bayesian Uncertainty Modeling}
In this section, we show how aleatoric and epistemic uncertainty can be jointly modeled using a single framework.

\subsection{Aleatoric Modeling}
Aleatoric uncertainty comes from inherent noise in the observations due to the probabilistic variability. To model this type of uncertainty during training, the network incorporates noise parameters ($\mu_t$, $\Sigma_{y_t}$) at time $t$, where $\mu$ denotes the mean and $\Sigma_{y_t}$ denotes the co-variance matrix for the ground-truth label $y_t$. The co-variance matrix $\Sigma_{y_t}$ is learned using negative log-likelihood loss function as follows:
\begin{equation}
\begin{split}
    \mathcal{L}_{A} &= -{\frac{1}{T}{  \sum_{t=T_{obs}+1}^{T_{pred}}\log(P(y_t|\mu_t,\Sigma_{y_t}))}}\\
    &= {\frac{1}{2T}{\sum_{t=T_{obs}+1}^{T_{pred}} {\frac{\norm{y_t-\mu_t}^2}{\Sigma_{y_t}}}}+log\ \Sigma_{y_t}}.
    \label{eq:1}
\end{split}
\end{equation}
We predict ($\mu_t$, $\Sigma_{y_t}$) at $T$ observed time-steps from time $T_{obs}+1$ to $T_{pred}$. Eq.~\ref{eq:1} is used to compute how likely the observations come from the posterior distribution $\mathcal{N}(\mu_t, \Sigma_{y_t})$. For numerical stability, having zeros in denominator is not suggested. Thus, we substitute log($\Sigma_{y_t}$) with $s_{yt}$, which results in Eq.~\ref{eq:2} as follows:
\begin{equation}
\begin{split}
    \Sigma_{y_t}&=\exp{(s_{yt})},\\
    \mathcal{L}_{A} &= {\frac{1}{2T}{\sum_{t=T_{obs}+1}^{T_{pred}} {\exp{(-s_{yt})}\norm{y_t-\mu_t}^2}}+s_{yt}}.
    \label{eq:2}
\end{split}
\end{equation}
\subsection{Epistemic Modeling}
Epistemic uncertainty is caused by the model's weight parameters that are inadequately measured from the observations. Thus, this type of uncertainty can be reduced by taking more measurements. Dropout is well-known in deep learning community, which is originally used as a regularization method to avoid over-fitting. However, a recent study in~\cite{gal2016dropout} introduced dropout to learn a distribution of weights to approximate variational inference in Bayesian modeling~\cite{variational_inference}. Given the dataset $X$, $Y$ the posterior over weights $P(w | X,Y)$ is approximated using a dropout distribution $q(w)$~\cite{yaringal_thesis}. During inference, we generate $N$ samples from the distribution $q(w)$ of the network's learned weight parameters $w$ using dropout. Then, $N$ number of noisy outputs are used to compute the variance $\Sigma_y$ between the predicted outputs $f^{\hat{w_i}}(x)$ and ground-truth labels $y_t$ at each time-step $t$. The details are shown in Eq~\ref{eq:3} as follows:
\begin{equation}
\begin{split}
    \mathcal{L}_E(w,P) &= -{\frac{1}{T}{\sum_{t=T_{obs}+1}^{T_{pred}}\log(P(y_t|f^{\hat{w}}(x_t)))}},\\
    \mu_y &= {\frac{1}{N}{\sum_{i=1}^{N}f^{\hat{w_i}}(x)}}\quad \hat{w}\sim q(w),\\
    \Sigma_y&={\frac{1}{N}{\sum_{i=1}^{N}f^{\hat{w_i}}(x)^Tf^{\hat{w_i}}(x)-\mu_y^T\mu_y}}.
    \label{eq:3}
\end{split}
\end{equation}
Note that the computation of the mean and variance is performed during inference using dropout. 
\subsection{Joint Modeling of Aleatoric and Epistemic Uncertainty}
\begin{figure*}[t]
    \centering
    \includegraphics[width=0.84\textwidth]{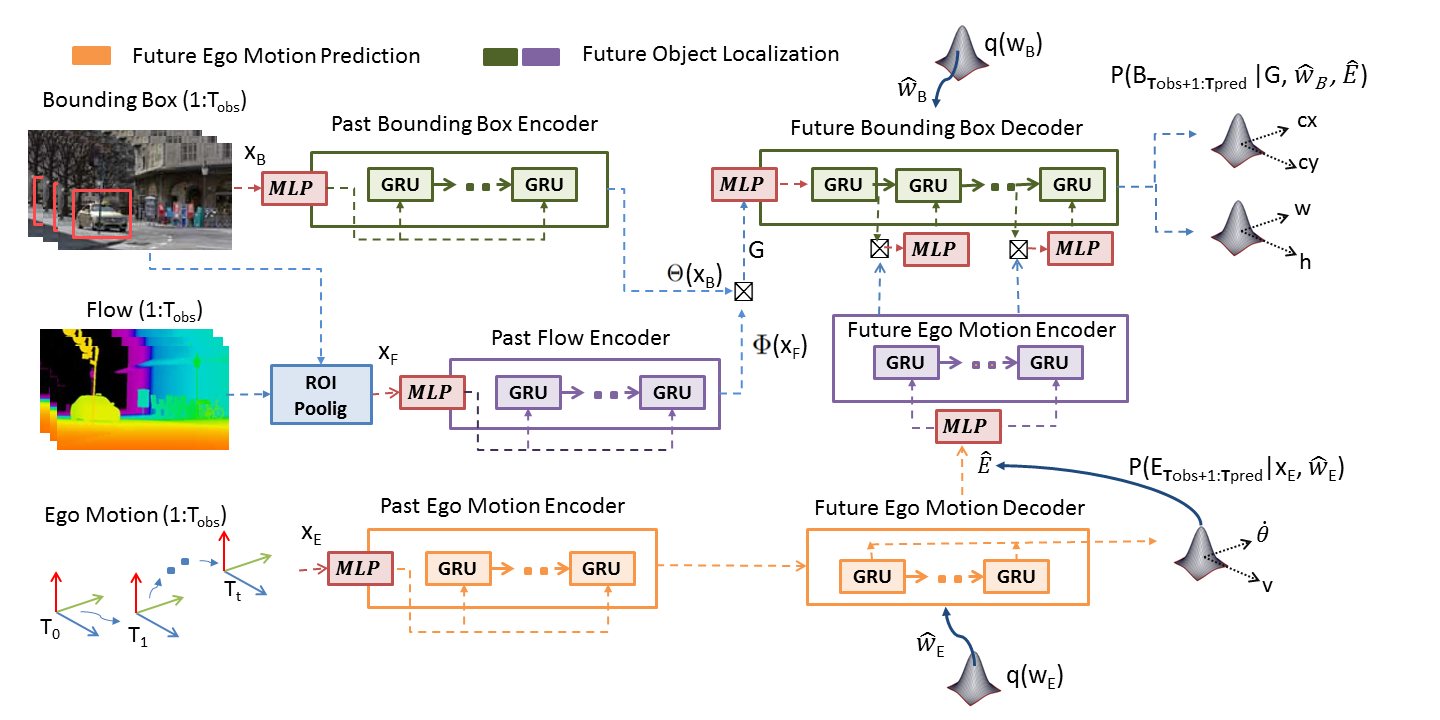}
    \caption{The proposed NEMO framework. 
    The \textit{Future ego-motion prediction} stream models the uncertainty of future ego-behavior. The \textit{Future object localization} stream encodes past bounding box and flow information to predict future motion of the target agent conditioned on the sampled future ego-motion. The resulting distribution is multi-modal and uncertainty-aware. 
    $\boxtimes$ is a concatenation operator.}
    \vspace{-0.51cm}
    \label{fig:block_diagram}
\end{figure*}
We update the noise parameters ($\mu_y$, $\Sigma_y$) by adding aleartoric uncertainty given in Eq.~\ref{eq:2} to epistemic uncertainty in Eq.~\ref{eq:3}. The total variance and mean is computed as shown in Eq.~\ref{eq:4}. $\{\hat{y_i},\hat{\Sigma_i}^2\}_{i=1}^N$ are set of $N$ sampled outputs from $f^{\hat{w_i}}(x)$ for randomly sampled weights $\hat{w}$ from the dropout distribution $q(w)$.
\begin{equation}
\begin{split}
    \hat{y_i},\hat{\Sigma_i} &= f^{\hat{w_i}}(x),\qquad \hat{w}\sim q(w)\\
    {\mu_y} &= {\frac{1}{N}{\sum_{i=1}^{N}\hat{y_i}}},\\
    {\Sigma_y} &=  {\frac{1}{N}} { \sum_{i=1}^{N} \hat{y_i}^T \hat{y_i} - \mu_y^T\mu_y }+{\frac{1}{N}} {\sum_{i=1}^{N} \hat{\Sigma_i}}.
    \label{eq:4}
\end{split}
\end{equation}
As a result, we output the noise parameters for the data posterior distribution $\mathcal{N}(\mu_t, \Sigma_{y_t})$ together with the learned distribution of the model's weights $q(w)$ during inference.
In practice, different node connections $\hat{w_i}$ are sampled for $N$ times using dropout, and corresponding aleatoric and epistemic uncertainty is computed using Eq.~\ref{eq:4}.

\section{NEMO Framework}
The proposed NEMO framework is designed to properly model the uncertainty of future ego-motion, which is most important to determine other agents' future motion in the egocentric view. As shown in Fig.~\ref{fig:block_diagram}, we divide the future forecast problem into two tasks: future ego-motion prediction and future object localization. For future ego-motion prediction, we first encode the past motion of the ego-vehicle and generate its future motion through the ego-motion decoder. To model the joint uncertainty of ego-motion, the model weights $\hat{w}_E$ for the ego-motion decoder are drawn from the weight distribution $q(w_E)$. Here, we generate multiple modes of prediction over the uncertainty distribution $P(E|x_E,\hat{w}_E)$ over the velocity $v$ and yaw rate $\dot{\theta}$, where $E = \{v, \dot{\theta}\}$ is the future ego-motion and $x_E$ is past ego-motion. 

From the other stream, the motion of other agents are encoded using the bounding box encoder and flow encoder, respectively. We then concatenate the encoded result and use the bounding box decoder to learn the weight parameters $\hat{w}_B$. To properly model their future behavior with respect to the noisy ego-motion, we use the output of the future ego-motion prediction as a prior. In this way, the bounding box decoder reacts to each modality $\hat{E}$ of the ego-vehicle, while predicting other agents' future motion. Similar to joint uncertainty modeling of the ego-motion decoder, the weights $\hat{w}_B$ for the bounding box decoder are drawn from the weight distribution $q(w_B)$. We estimate the noise parameters for the center $(c_x, c_y)$ and the dimension $(w, h)$ of the bounding box using the weights $\hat{w}_B$. Finally, we predict $B = \{c_x, c_y, w, h\}$
by sampling from the uncertainty distribution $P(B | G,\hat{w}_B,\hat{E})$, $G=\Phi(x_F) \boxtimes \Theta(x_B)$ concatenation of past flow $x_F$ and bounding box $x_B$ encoding ($\Phi$ is flow encoder, $\Theta$ is past bounding box encoder, and $\boxtimes$ is concatenation operator).
\subsection{Future Ego-Motion Prediction}
Given the past observations, predicting the future ego-motion must account for multiple possibilities. Thus, we model multi-modal predictions for the ego-motion with uncertainty estimates. We take the past observations $x_{E,t}$ from IMU odometry ($v_t$, $\dot{\theta_t}$) for $T_{obs}$ time steps and encode the ego-motion using Gated Recurrent Units (GRU). Multi Layer Perceptron (MLP) is used to convert the past ego-motion to the embedding of the GRU. 
The prediction output of a GRU-based decoder is a 5-dimensional vector $[\mu_v,\mu_{\dot{\theta}},\sigma_v,\sigma_{\dot{\theta}},\rho]$ at each future time step from $T_{obs}+1$ to $T_{pred}$, where $\mu_v$ is mean and $\sigma_v$ is noise in velocity prediction, $\mu_{\dot{\theta}}$ is mean and $\sigma_{\dot{\theta}}$ is noise in yaw rate prediction, and $\rho$ is correlation coefficient between those two dimensions. During inference, we sample velocity $\hat{v_t}$ and yaw rate $\hat{\dot{\theta_t}}$ from the uncertainty distribution generated by the noise parameters. 
The input is $x_{E,t}=[v_t,\dot{\theta_t}]_{t=\{1:T_{obs}\}}$ and output is $y_{E,t}=[\hat{v_t},\hat{\dot{\theta_t}}]_{t=\{T_{obs}+1:T_{pred}\}}$.

\subsection{Future Object Localization}
We use the past bounding box information $x_{B,t}$ and past ROI pooled Flow information $x_{F,t}$, which are separately processed using the respective GRU encoders. In addition, we use the predicted future ego-motion $y_{E,t}$ as a prior to generate future motion of the target $y_{O,t}$ at time $t$. The output of future object localization is a 10-dimensional vector $[\mu_{c_x},\mu_{c_y},\sigma_{c_x},\sigma_{c_y},\rho_{c},\mu_{w},\mu_{h},\sigma_{w},\sigma_{h},\rho_{d}]$ at each future time step, where $(\mu_{c_x},\mu_{c_y},\sigma_{c_x},\sigma_{c_y},\rho_{c})$ and 
$(\mu_{w},\mu_{h},\sigma_{w},\sigma_{h},\rho_{d})$
are a set of mean and co-variance parameters for the center and for the bounding box dimension, respectively. By assuming two 2D-Gaussian functions for the uncertainty, we reduced the number of parameters to regress from 20 (4D-Gaussian) to 10 (2 2D-Gaussian). The output center ($\hat{c}_{x,t}, \hat{c}_{y,t}$) and dimension ($\hat{w_t}, \hat{h_t}$) of the bounding boxes are sampled from the uncertainty distribution generated by these noise parameters.  
\section{Experiments}
\subsection{Dataset}
The HEV-I dataset \cite{brian_icra} is publicly available and consists of 2477 vehicles in 230 videos collected from urban driving scenarios. The dataset includes the motion of the ego-vehicle obtained by ORB-SLAM2~\cite{orbslam}. However, the estimated translation is a normalized unit vector which does not recover the full 3D motion of the ego-vehicle. Moreover, the dynamic motion of surrounding agents often causes association errors, which severely affect the rotation estimates. Therefore, we provide IMU odometry $(v,\dot{\theta})$ decoded from the CAN message of the ego-vehicle. We observed that a drift error is less than 0.2 $meters$ for new IMU odometry as compared to LIDAR odometry for the HEV-I sequences.

\begin{figure*}[t]
\begin{center}
  \mbox{
      \subfigure[ \label{fig:scene1a}]{\includegraphics[width=0.28\textwidth]{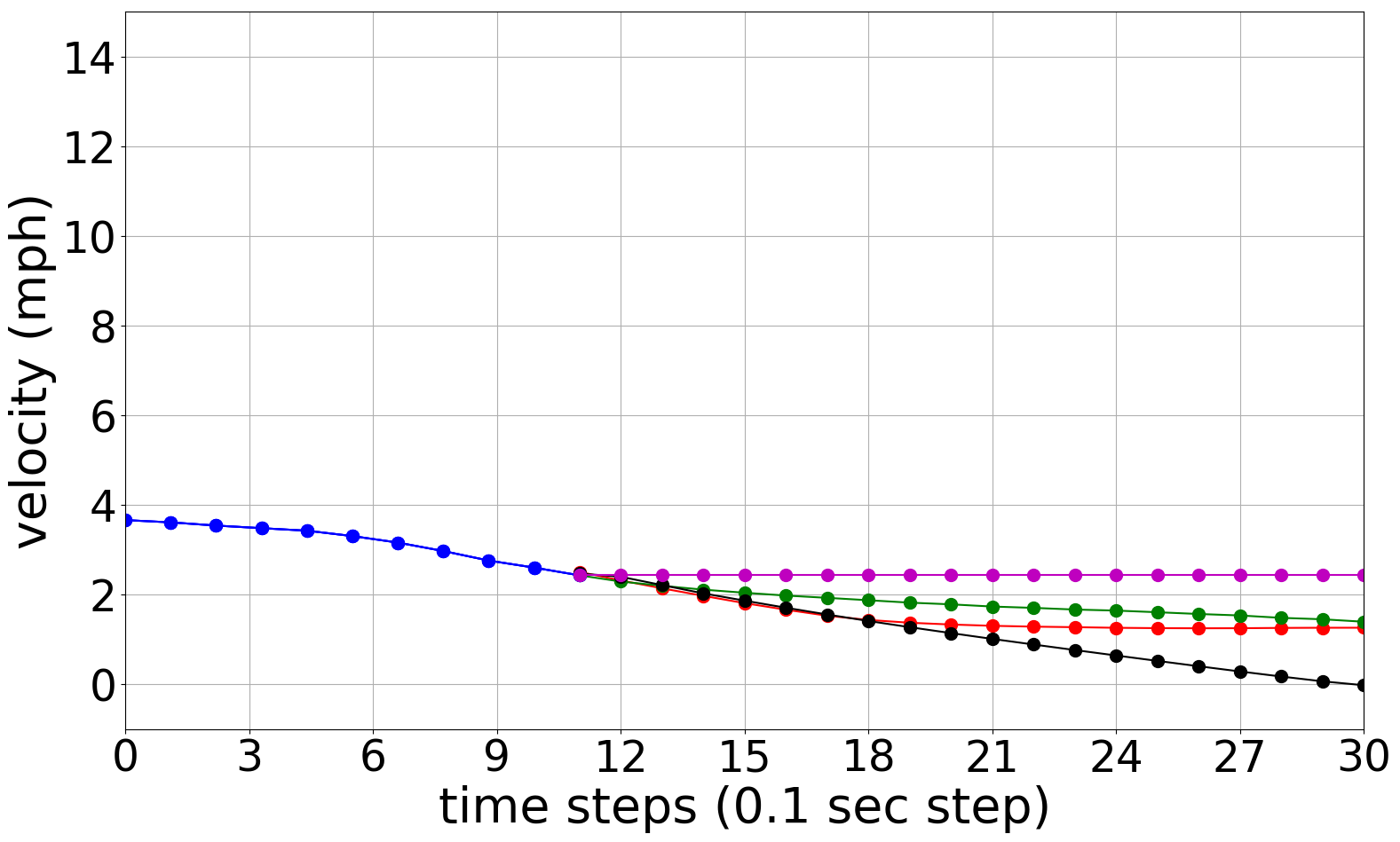}}\quad
      \subfigure[ \label{fig:scene1b}]{\includegraphics[width=0.28\textwidth]{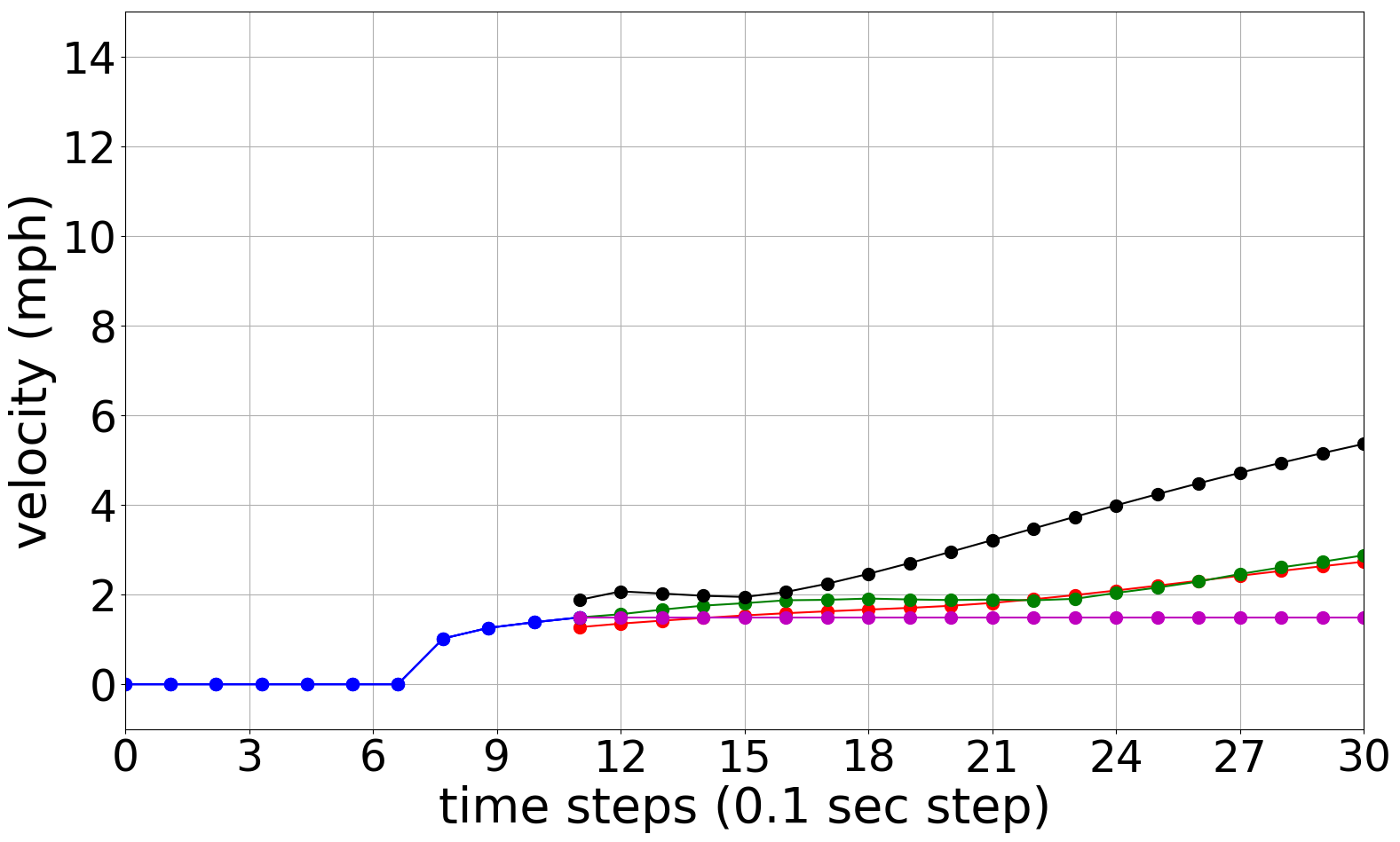}}
      \subfigure[ \label{fig:scene1c}]{\includegraphics[width=0.28\textwidth]{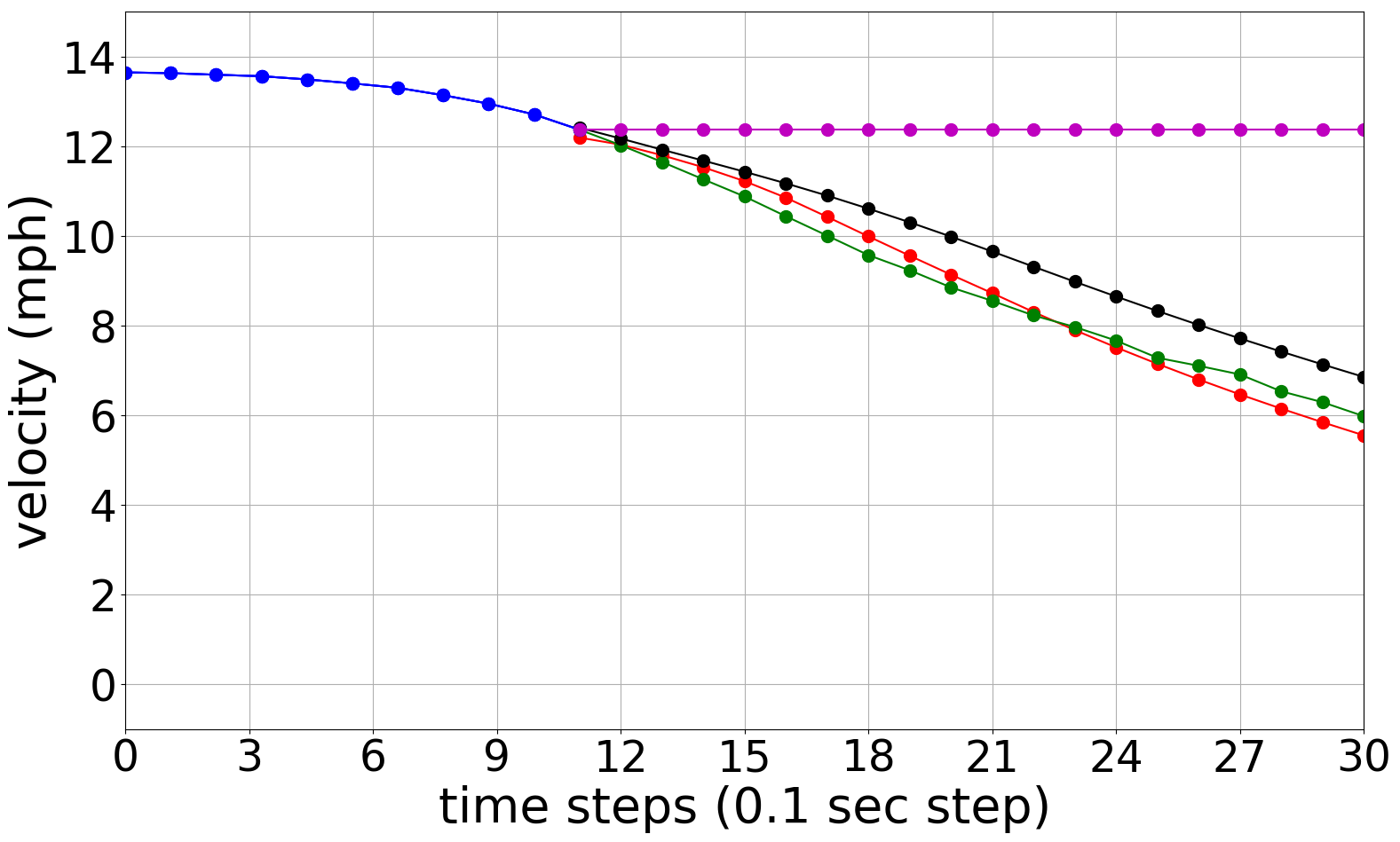}}
    }
  \mbox{
      \subfigure[ \label{fig:scene1a}]{\includegraphics[width=0.28\textwidth]{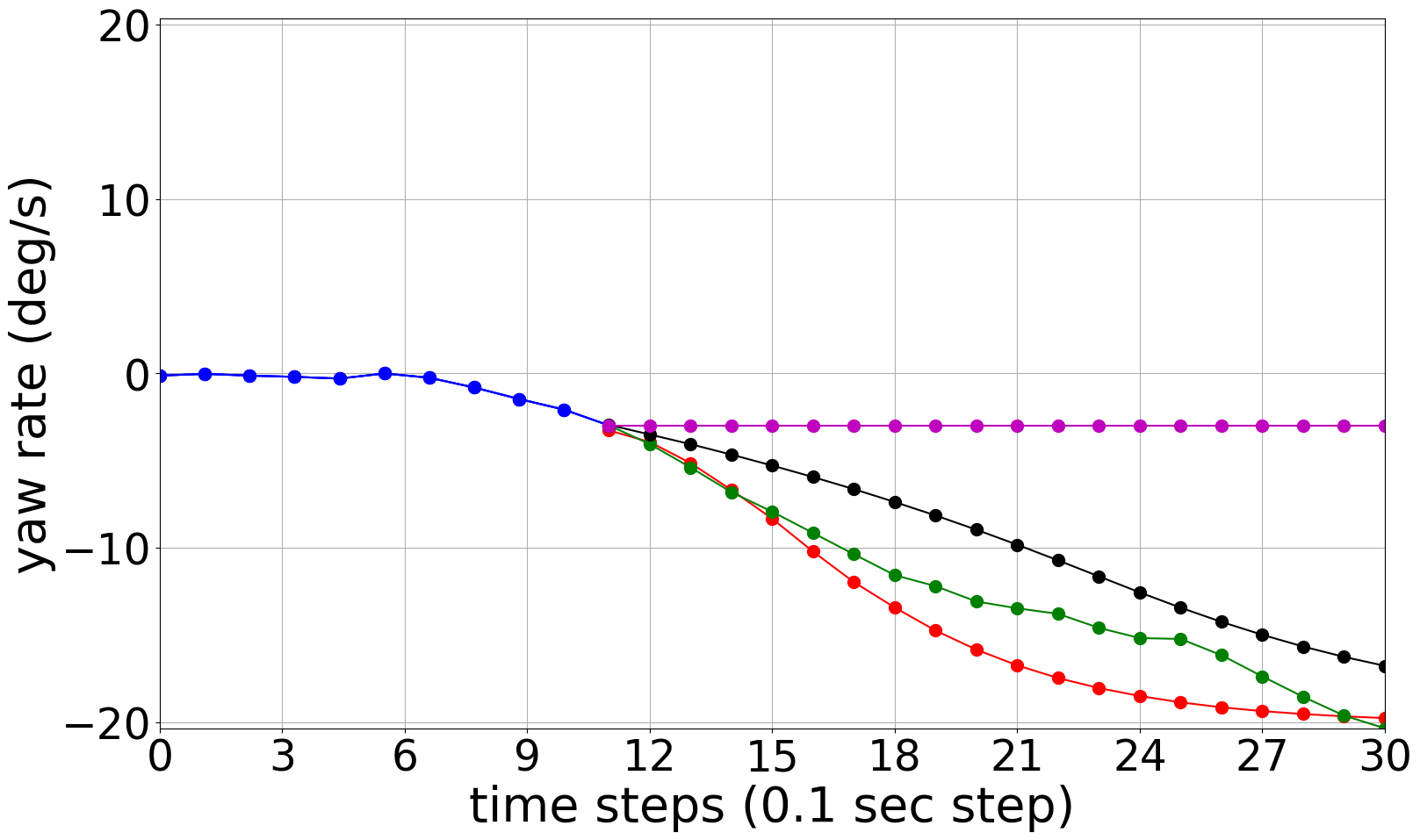}}\quad
      \subfigure[ \label{fig:scene1b}]{\includegraphics[width=0.28\textwidth]{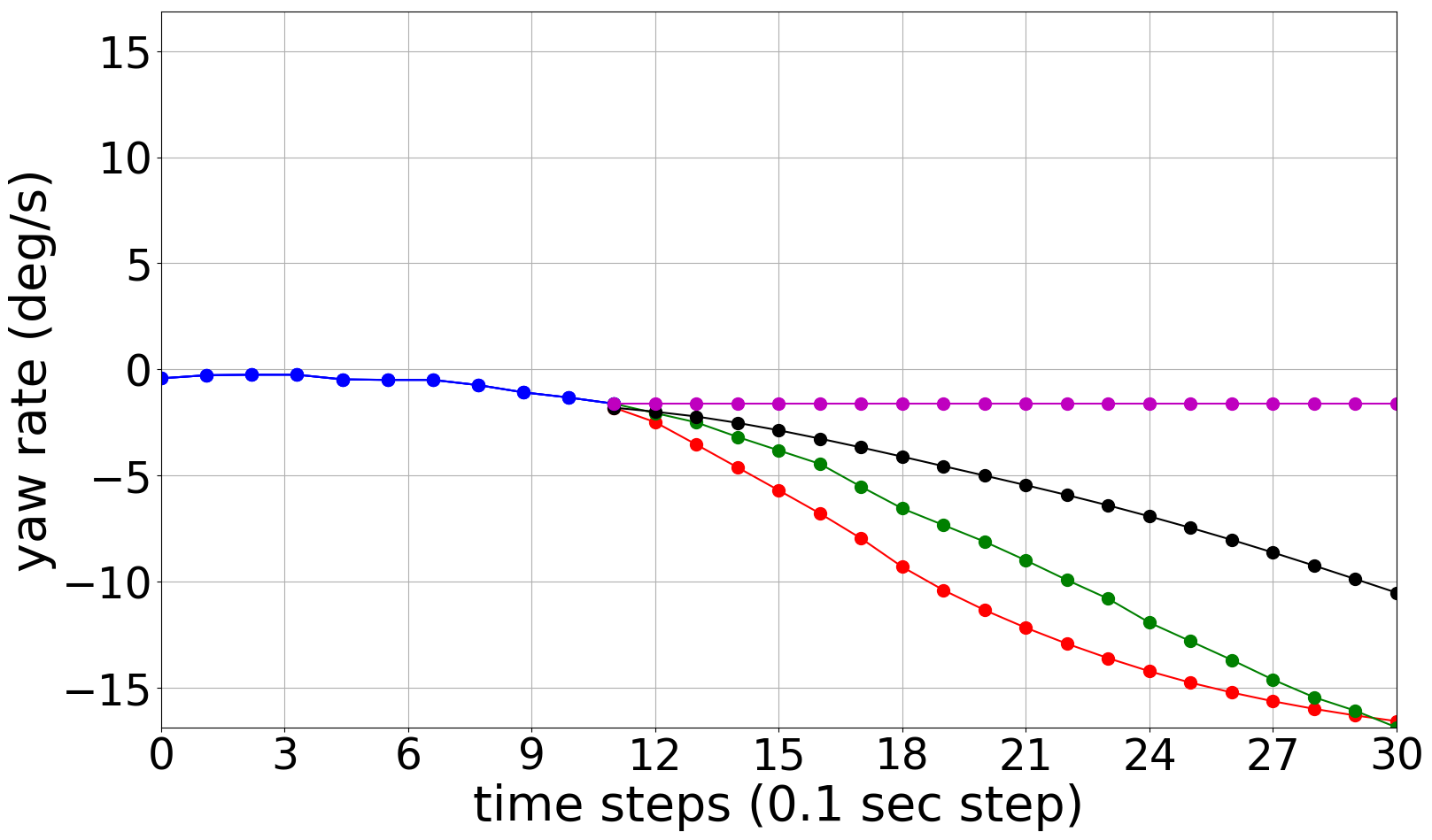}}
      \subfigure[ \label{fig:scene1c}]{\includegraphics[width=0.28\textwidth]{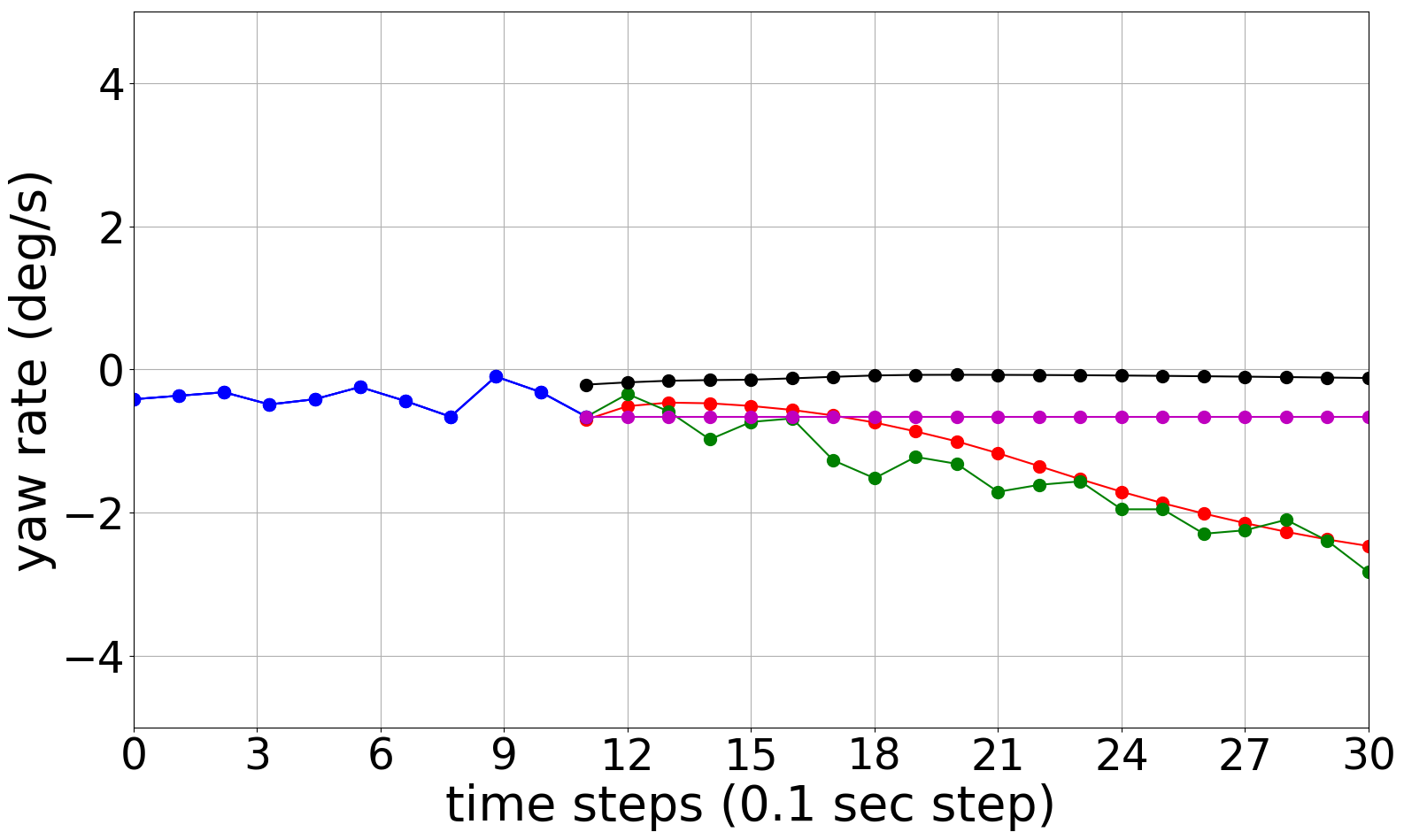}}
    }
    \end{center}\vspace{-0.51cm}
  \caption{Future ego-motion prediction. (a,b,c) velocity and (d,e,f) yaw rate. Given the past observation \protect\inlinegraphics{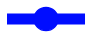} and future ground-truth \protect\inlinegraphics{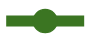}, Const-Vel \protect\inlinegraphics{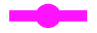} and RNN \protect\inlinegraphics{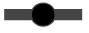} models are compared with RNN-AE (Ours) \protect\inlinegraphics{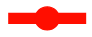}.
  }
  \label{fig:4}
   \vspace{-0.51cm}
\end{figure*}

\begin{figure*}[t]
\begin{center}
  \mbox{
      \subfigure[ \label{fig:scene1a_vel_ours}]{\includegraphics[width=0.28\textwidth]{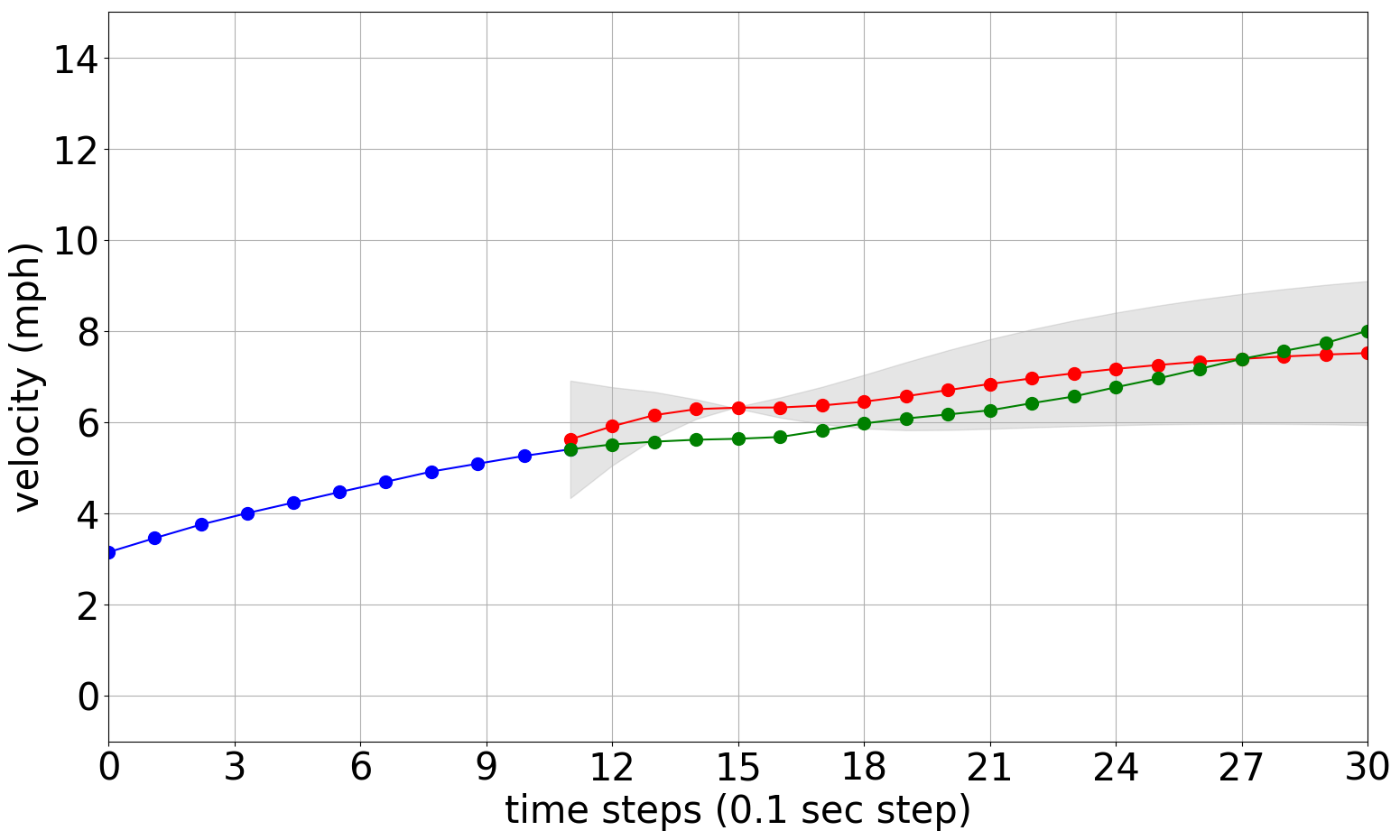}}\quad
      \subfigure[ \label{fig:scene1b_vel_ours}]{\includegraphics[width=0.28\textwidth]{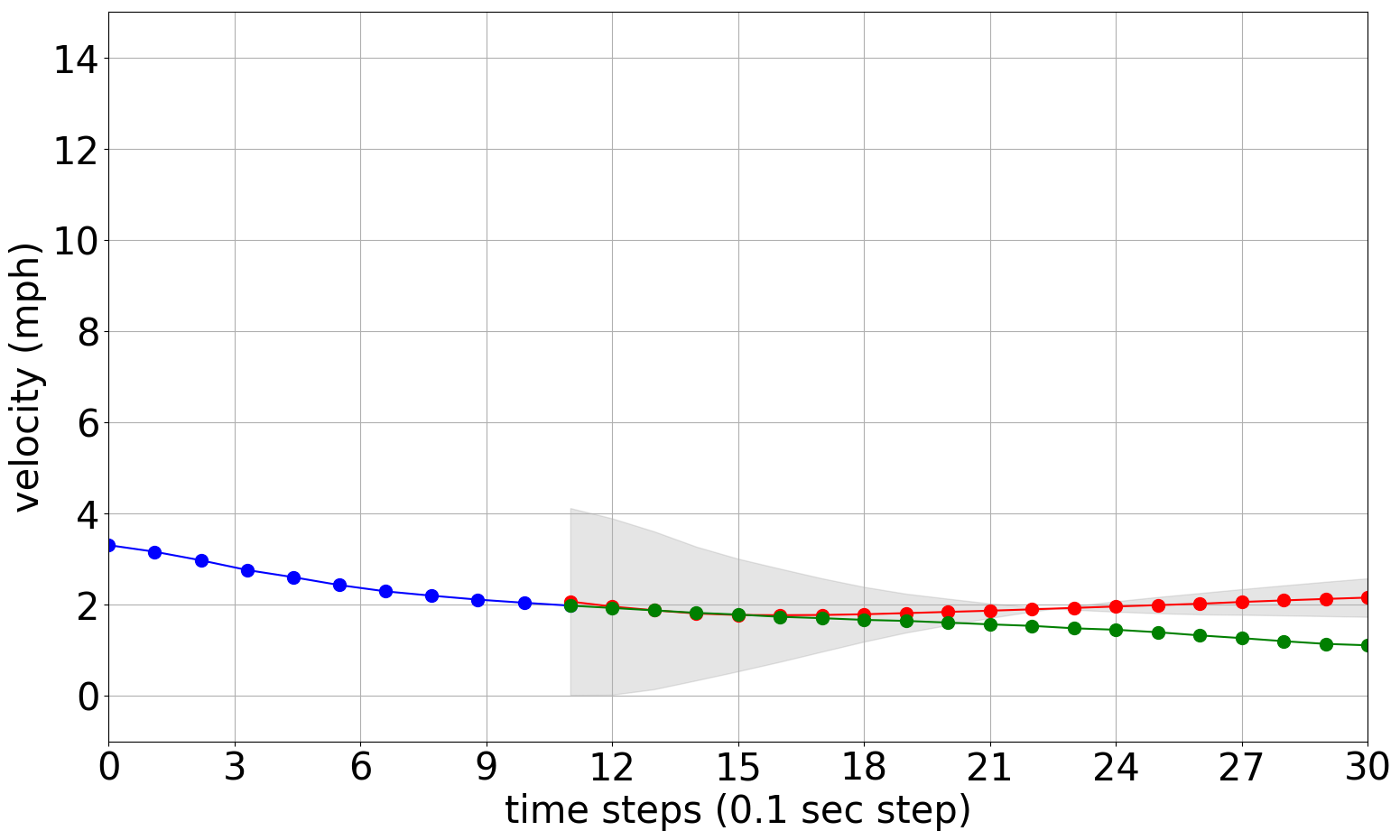}}
      \subfigure[ \label{fig:scene1c_vel_ours}]{\includegraphics[width=0.28\textwidth]{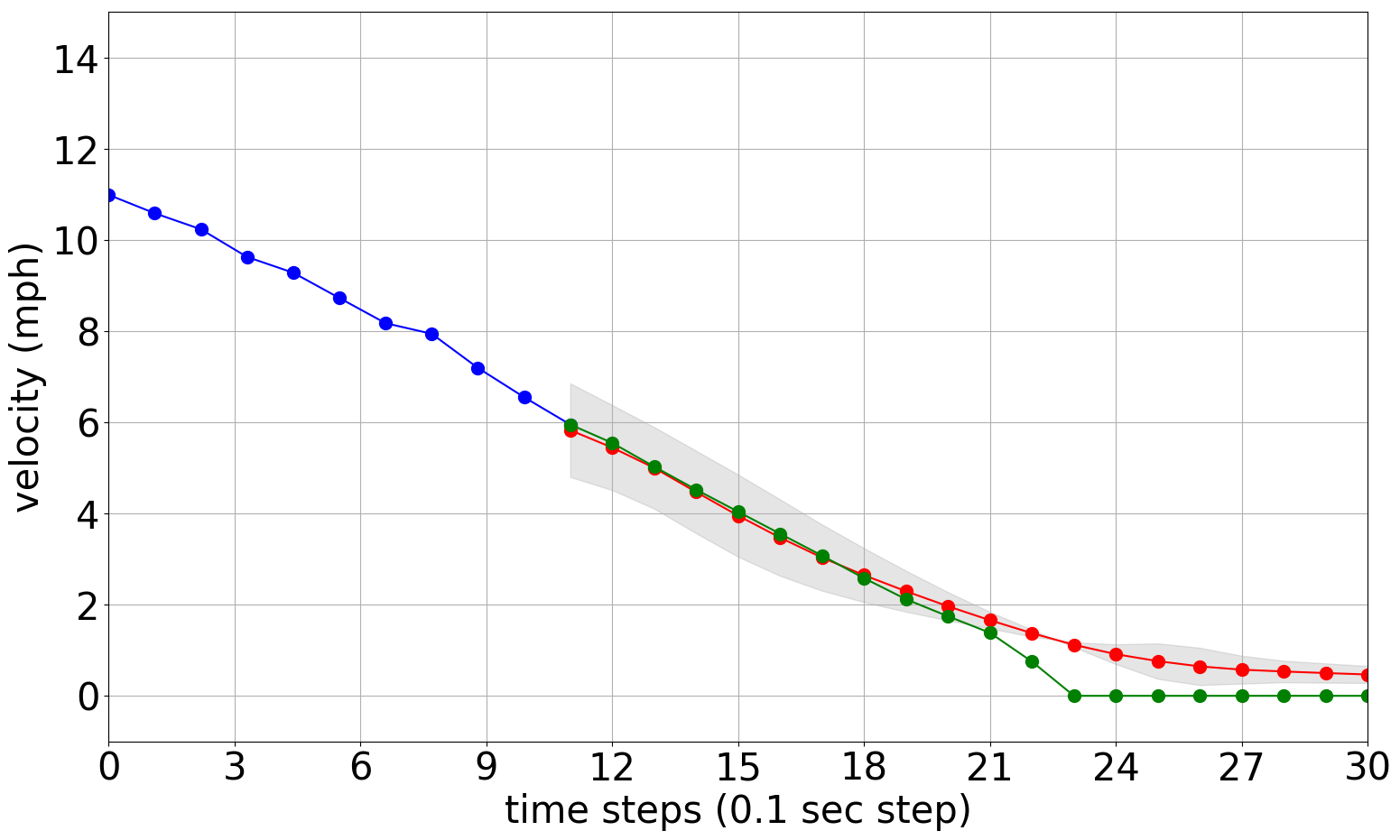}}
    }
  \mbox{
      \subfigure[ \label{fig:scene1a_yaw_ours}]{\includegraphics[width=0.28\textwidth]{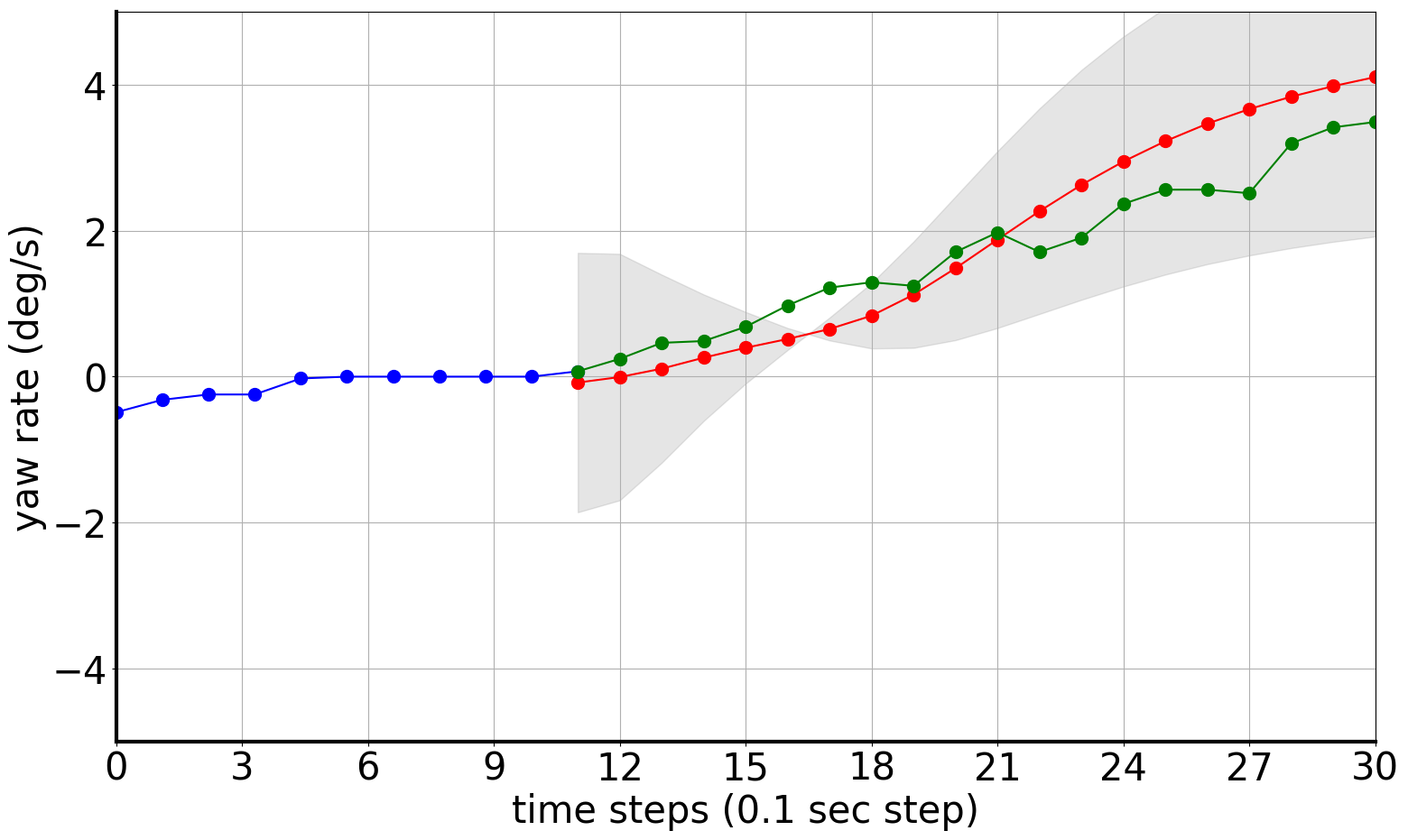}}\quad
      \subfigure[ \label{fig:scene1b_yaw_ours}]{\includegraphics[width=0.28\textwidth]{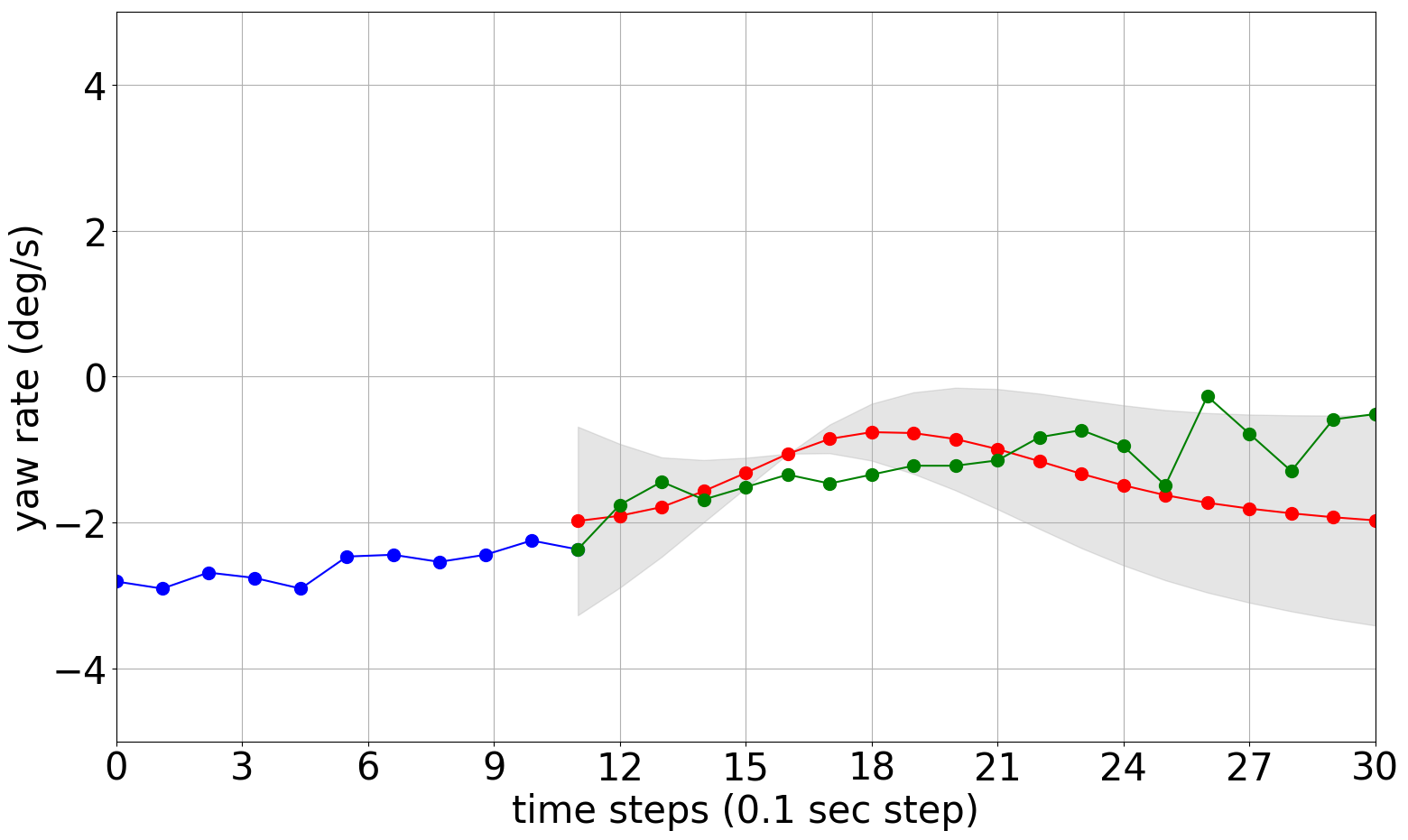}}
      \subfigure[ \label{fig:scene1c_yaw_ours}]{\includegraphics[width=0.28\textwidth]{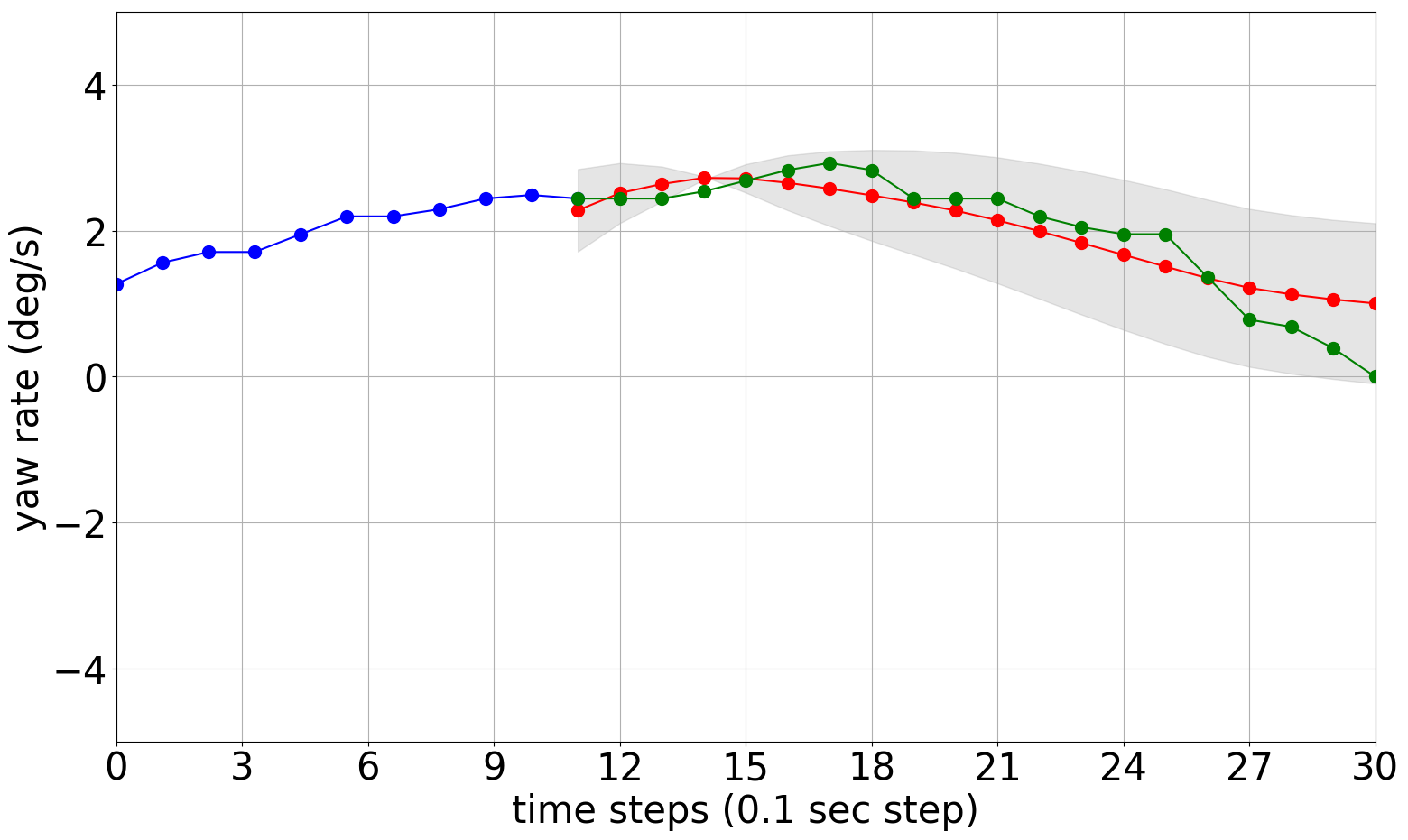}}
    }
    \end{center}\vspace{-0.51cm}
  \caption{Future ego-motion prediction using NEMO (RNN-AE) with the uncertainty. (a,b,c) velocity and (d,e,f) yaw rate of the ground-truth \protect\inlinegraphics{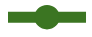} and RNN-AE (Ours) \protect\inlinegraphics{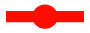} is plotted with the uncertainty \protect\inlinegraphics{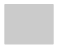} at each time step.
}
  \label{fig:6}
   \vspace{-0.51cm}
\end{figure*}
\subsection{Implementation}
NEMO is trained with a TITAN Xp GPU using the PyTorch framework. We first train the ego-motion prediction stream from scratch. Then, the learned model is jointly optimized with the future object localization stream. 

\subsubsection{Future Ego Motion Prediction}
We use a batch size of 32 and learning rate of 0.001 for negative log-likelihood loss as future ego motion loss 
with the RMSProp optimizer. For the learning rate, we drop the value by a factor of 2 after every 20 epochs. The network converges after 100 epochs. For evaluation, we reconstruct the 2D trajectory from the predicted velocity and yaw rate using Eq.~\ref{eq:7} with respect to the last observed frame, assuming planar motion.
\begin{equation}
\begin{split}
    R_{0}^i &= \prod_{t=0}^{i-1}R_t^{t+1},\\
    T_{0}^i &= T_{0}^{i-1}+R_0^{i-1}T_{i-1}^i,
    \label{eq:7}
\end{split}
\end{equation}
where $R_{i}^{i+1}\in \mathbb{R}^{2\times2}$ is a 2D rotation matrix and $T_{i}^{i+1}\in \mathbb{R}^{2}$ is a 2D translation vector. We use a right handed coordinate system for the ego-motion. Note that velocity and yaw rate is converted into translation in meters and degrees between every time step using the time interval of 0.1 sec.

\subsubsection{Future Object Localization}
We use input image of size $W = 1920$ and $H=1200$ $pixels$. The bounding box centers and dimensions are normalized to a range of $[0,1]$. 
While training the module, we use a batch size of 32 and learning rate of 0.001 that is reduced by a factor of 5 after every 20 epochs. We use a weighting $\lambda_{e} = 0.2$ for the pre-trained model for future ego motion prediction loss 
and $\lambda_{f} = 1$ for future object localization loss. 
\section{Results}
Prior works \cite{cvpr_uncertainty,brian_icra} report that a 1 second prediction time is sufficient for safe operation of the vehicle travelling with a speed up to 25 MPH.   However, for natural driving in urban areas, this is an underestimate since we found that vehicles travel with a speed up to 43 MPH in the HEV-I dataset. Thus, we observe the past 1 second, and make predictions 2 seconds in the future. We sample $k=10$ future predictions from the distribution and report the result with a minimum error as $y_{opt} = \operatorname*{min}_{k} {\norm{\hat{y}^{k}-y}_2}$, where $\hat{y}^k$ is trajectory prediction sample and $y$ is ground truth trajectory. For evaluation, we compute the Average Distance Error (ADE) and Final Distance Error (FDE) for motion prediction, and Final Intersection over Union (FIOU) for bounding box prediction. The reported ADE/FDE for ego-motion prediction is in units of $meters$, while those for bounding box prediction is in $pixel$ units. 
\subsection{Future Ego Motion Prediction}
\begin{figure*}[t]
\begin{center}
  \mbox{
      \subfigure{\includegraphics[width=0.28\textwidth]{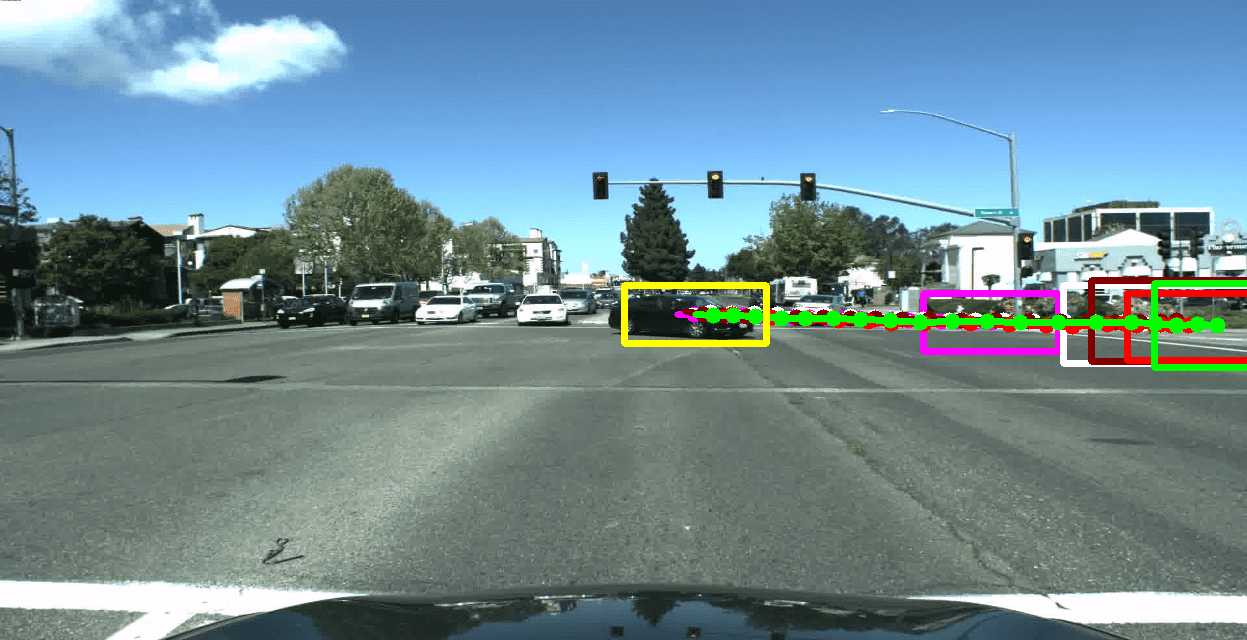}}\quad
      \subfigure{\includegraphics[width=0.28\textwidth]{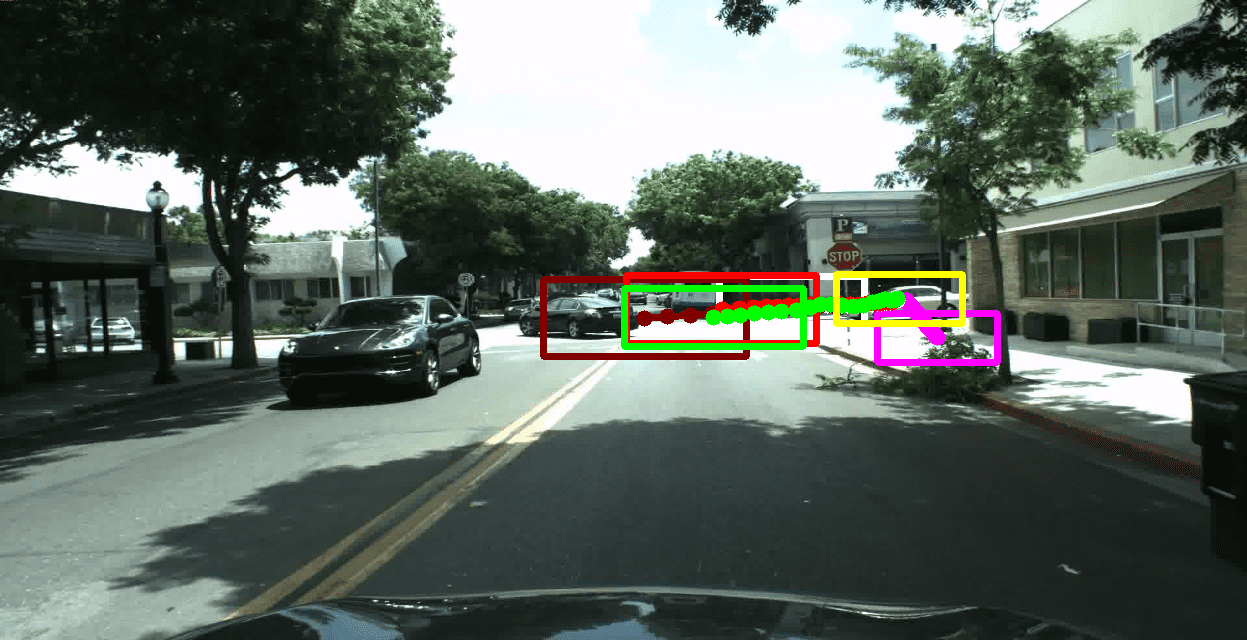}}
      \subfigure{\includegraphics[width=0.28\textwidth]{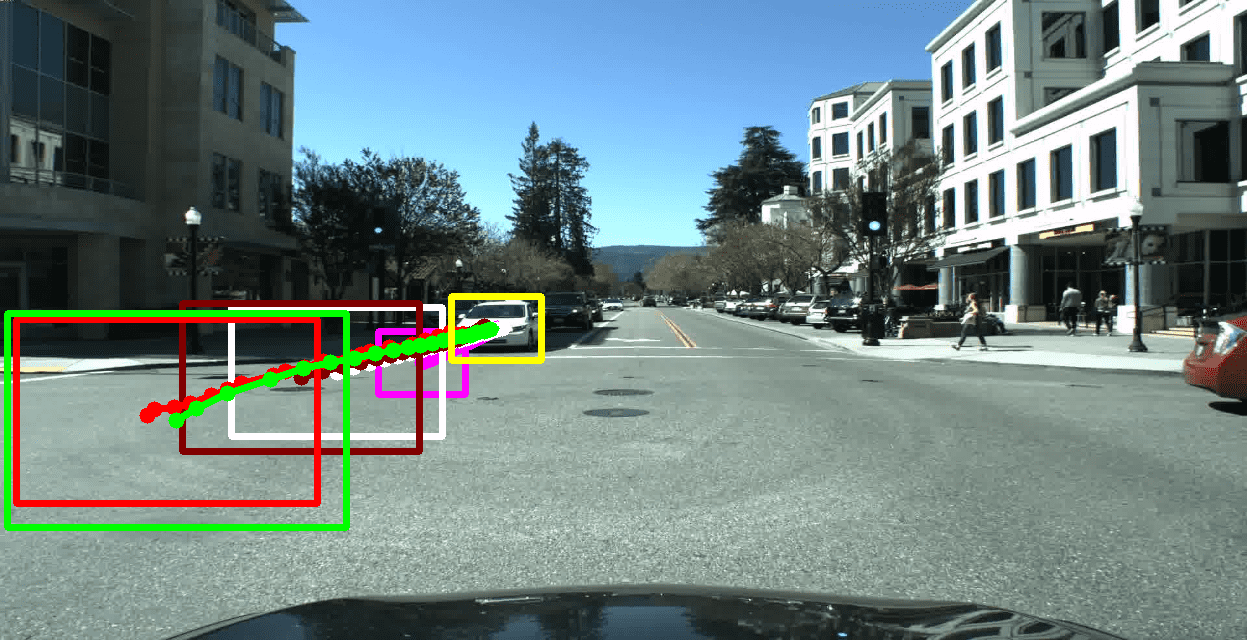}}
    }
    \end{center}
  \caption{Example scenarios for future object localization. Given the last observation \protect\inlinegraphics{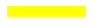} at time $t=T_{obs}$ and the future ground-truth \protect\inlinegraphics{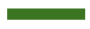} at time $t=T_{pred}$, Const-Vel \protect\inlinegraphics{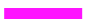}, RNN-P (ORB) \protect\inlinegraphics{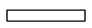}, and RNN-P (IMU) \protect\inlinegraphics{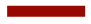} models are compared to RNN-AE (Ours) \protect\inlinegraphics{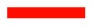}. Also, the predicted trajectory of RNN-AE (Ours) \protect\inlinegraphics{figures/comp_markers/m2.png} is visualize with the ground-truth \protect\inlinegraphics{figures/comp_markers/m3.png}.
  }
  \label{fig:3}
  \vspace{-0.51cm}
\end{figure*}

\begin{figure*}[t]
\begin{center}
  \mbox{
      \subfigure{\includegraphics[width=0.28\textwidth]{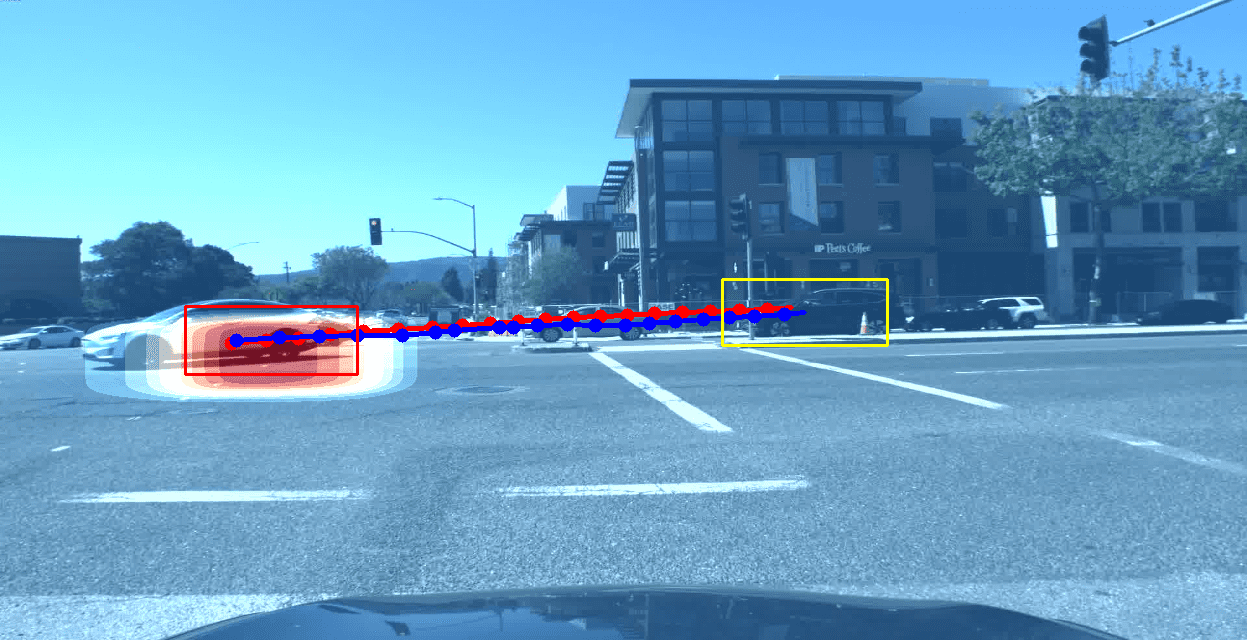}}\quad
      \subfigure{\includegraphics[width=0.28\textwidth]{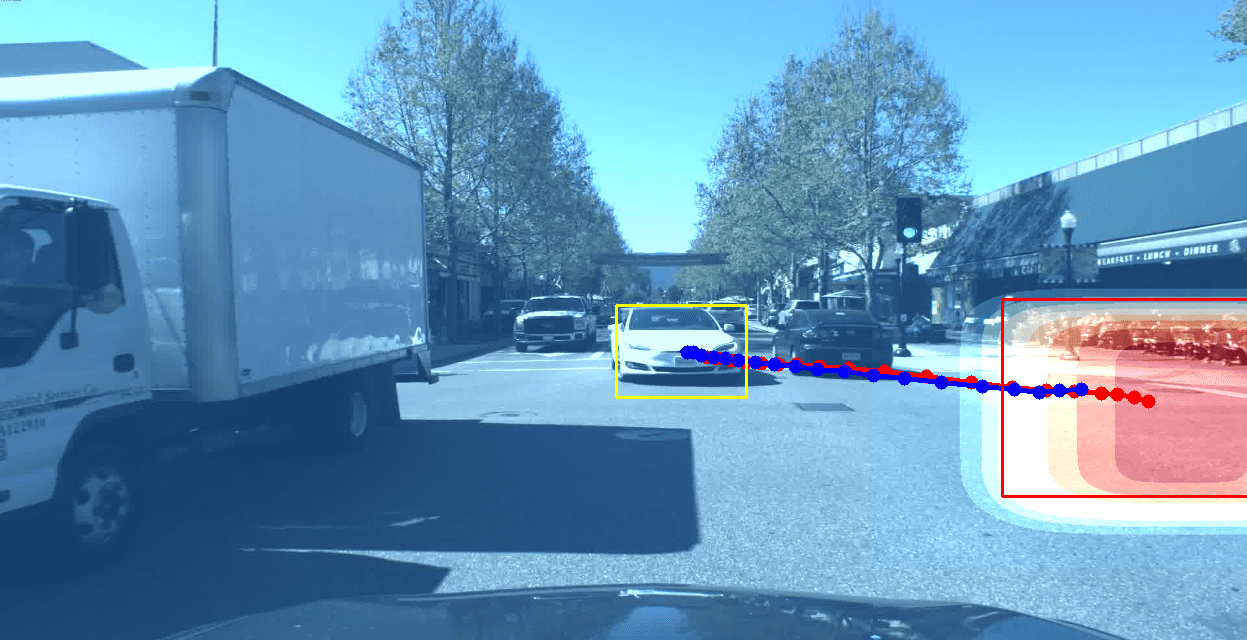}}
      \subfigure{\includegraphics[width=0.28\textwidth]{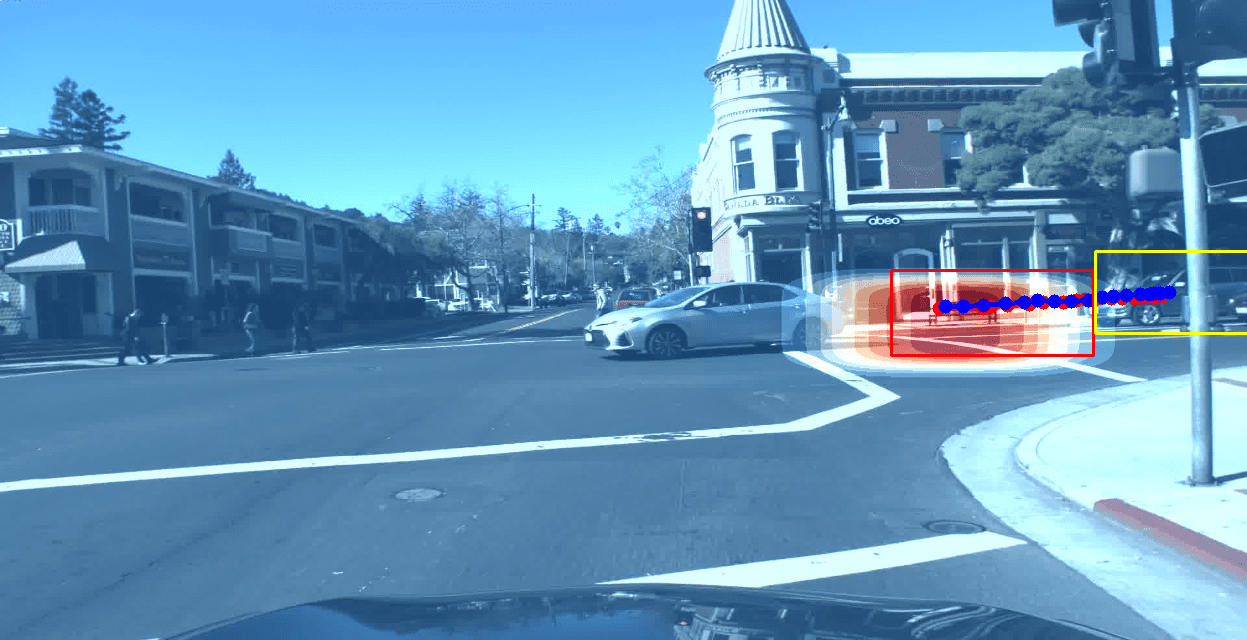}}
      \subfigure{\includegraphics[width=0.035\textwidth]{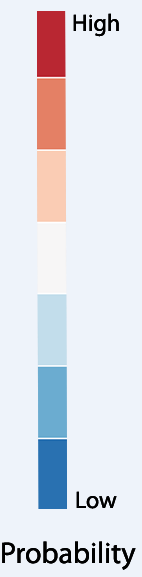}}
    }
    \end{center}
  \caption{Qualitative evaluation of NEMO (RNN-AE) with the uncertainty of future object localization. The predicted centers of the bounding box from $T_{obs}+1$ to $T_{pred}$ are shown as a trajectory \protect\inlinegraphics{figures/markers_ours/m2.png} with ground-truth \protect\inlinegraphics{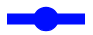}. Also, the bounding box \protect\inlinegraphics{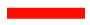} at $T_{pred}$ is sampled from the probability distribution (red indicates high probability). 
  }
  \label{fig:5}
   \vspace{-0.51cm}
\end{figure*}

We use Const-Vel \cite{constvel} as one of our baselines where the output for future 20 time steps is the same as the input observed at time $t=10$. The RNN baseline has a GRU-based encoder and decoder, which highly improves the performance compared to the Const-Vel baseline. For RNN-E, we model the epistemic uncertainty with a dropout in the decoder's MLP layer. 
RNN-A is with aleatoric uncertainty, where we sample 10 trajectories from the learned likelihood distribution parameters. RNN-AE is a combined aleatoric and epistemic uncertainty modeling. We observe that uncertainty modeling improves performance in all three cases (RNN-E, RNN-A, RNN-AE) compared to their counterparts (Const-Vel, RNN). Overall, RNN-AE performs better than all the baselines as shown in Fig.~\ref{fig:4} and Tbl.~\ref{tbl:1}, validating the efficacy of uncertainty modeling. The uncertainty estimates are shown in Fig.~\ref{fig:6}. 

\begin{table}[h!]
\centering
\begin{tabular}{|l|l|l|}
\hline
Method & ADE $\downarrow$ & FDE $\downarrow$   \\\hline
Const-Vel~\cite{constvel} & 0.3089 & 0.8386   \\ \hline
RNN & 0.1824 & 0.4275  \\ \hline
RNN-E & 0.1530 & 0.3907 \\ \hline
RNN-A & 0.1501 & 0.3279\\ \hline    
RNN-AE (Ours) &\textbf{0.1324} & \textbf{0.3031} \\ \hline
\end{tabular}
 \caption{Quantitative results for future ego-motion prediction. ADE/FDE errors are reported in $meters$.}
\vspace{-0.51cm}
\label{tbl:1}
\end{table}
\subsection{Future Object Localization}
We use the Const-Vel baseline for bounding box prediction in pixel coordinates. However, this baseline does not consider the scaling factors in the egocentric videos. For fair comparison, we linearly scale the bounding box dimensions using the transformation of the last two observations. As shown in Tbl~\ref{tbl:2}, linearly scaled bounding box dimensions improves the FIOU performance. 
RNN-NP does not use any priors of the ego-motion. Interestingly, we observe improvement in both ADE and FDE,  but its FIOU is degraded as compared to the Const-Vel baseline. RNN-P (ORB) uses ORB-SLAM2 \cite{orbslam} based ego-motion as a prior \cite{brian_icra}, while RNN-P (IMU) uses IMU odometry for ego-motion prediction similar to~\cite{cvpr_uncertainty}. We observe that IMU-based ego-motion, RNN-P (IMU), improves the performance when compared with the ORB-SLAM2-based ego-motion, which validates our claim to use IMU odometry for future object localization. 
For RNN-AP, we use the pre-trained ego-motion prediction module with aleatoric uncertainty and train jointly with future object localization. Similarly, RNN-EP uses the pre-trained ego-motion prediction module with epistemic uncertainty and is trained jointly with future object localization. Use of these uncertainty models to condition other agents' motion forecast significantly improves the overall performance. This comparison validates the rationale of our use of the uncertainty to model more robust interactions of other agents with the ego-vehicle. For RNN-A, both ego-motion prediction and future object localization modules are trained with aleatoric uncertainty. Similarly, RNN-E is trained with epistemic uncertainty for both tasks. These baseline models further decrease the error rate compared to RNN-AP and RNN-EP. 
Finally, RNN-AE (Ours) models aleatoric and epistemic uncertainty throughout the NEMO pipeline, which has the best performance in predicting future motion of agents as well as others' bounding box locations and scales. Fig.~\ref{fig:3} qualitatively evaluates how NEMO (RNN-AE) performs against other methods, and Fig.~\ref{fig:5} visualizes the uncertainty of future object localization. From these results, we conclude that NEMO properly captures the interactive behaviors of road agents with respect to the ego-vehicle with the uncertainty of future forecast in the egocentric view.
\begin{table}[h!]
\centering
\begin{tabular}{|l|l|l|l|}
\hline
Method & ADE $\downarrow$ & FDE $\downarrow$ & FIOU $\uparrow$  \\\hline
Const-Vel (w/o scaling)~\cite{constvel} & 92.27 & 203.84 &0.2660 \\ \hline
Const-Vel (w/ scaling)~\cite{constvel} & 92.27 & 203.84 &0.2999 \\\hline
RNN-NP &70.97 & 146.23&0.2703 \\ \hline
RNN-P (ORB)~\cite{brian_icra} &59.06&123.81& 0.2985\\ \hline
RNN-P (IMU)~\cite{cvpr_uncertainty} & 54.81 & 113.35  &0.3544\\ \hline
RNN-AP &51.00  & 107.2 & 0.4009\\ \hline    
RNN-EP &51.81 & 108.7 & 0.4282\\ \hline
RNN-A & 49.86 & 105.3 & 0.4652\\ \hline
RNN-E & 49.91& 106.02&0.4803\\ \hline
RNN-AE (Ours)& \textbf{49.02}  &\textbf{100.26} &\textbf{0.5194}\\ \hline
\end{tabular}
 \caption{Quantitative results for future object localization. ADE/FDE are reported in $pixel$ on an image of size 1200x1920.}
 \vspace{-0.51cm}
 \label{tbl:2}
\end{table}

\begin{figure*}[!t]
\begin{center}
  \mbox{
      \subfigure[]{\includegraphics[width=0.42\textwidth]{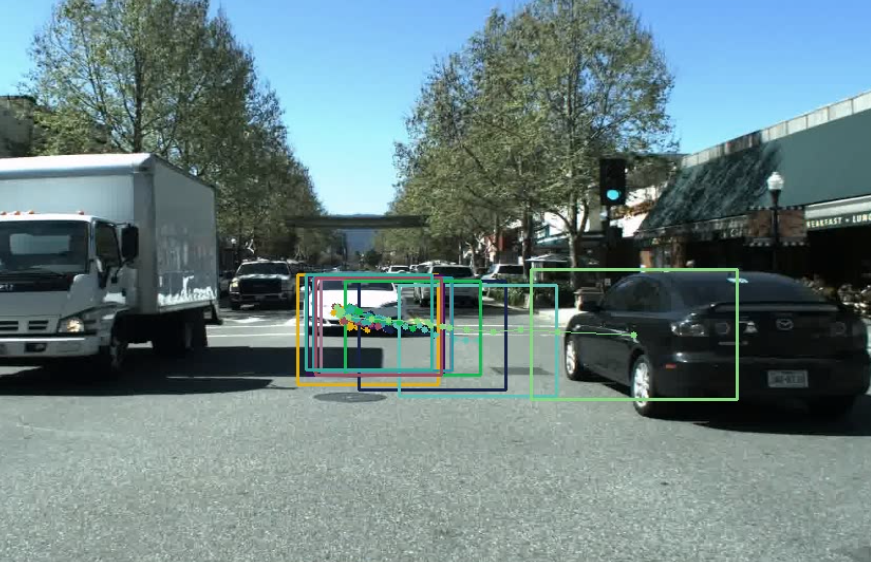}\label{fig:7a}}\quad
      \subfigure[]{\includegraphics[width=0.46\textwidth]{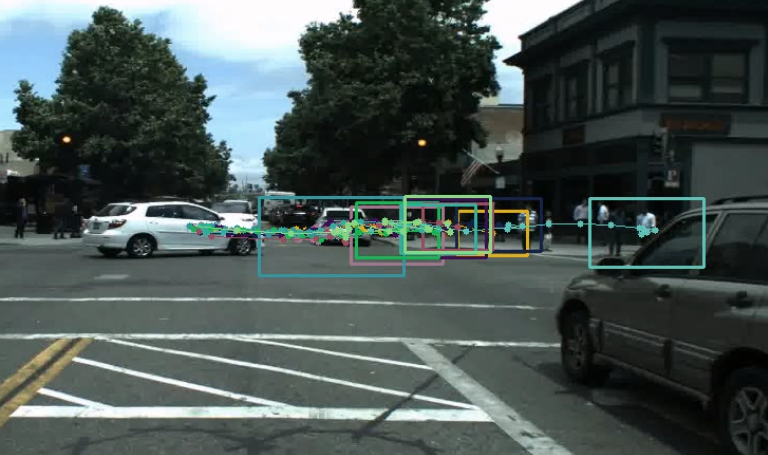}\label{fig:7b}}
    }
    \end{center}
  \caption{Example scenarios of multi modality for future object localization.
  }
  \label{fig:7}
  \vspace{-0.51cm}
\end{figure*}

\subsection{Multi Modality}
We further evaluate on the multi-modal capability of our framework in Fig.~\ref{fig:7}. To generate multi-modal future trajectories, we sample 10 future positions from the prediction distribution at each time-step and condition the next time-step prediction on current time-step output. The resulting diverse trajectories at intersection scenarios are shown with different colors. The white vehicle in Fig.~\ref{fig:7a} shows multiple possible motions. Based on the uncertain ego-future, it may turn to its left when the ego-car is turning left (light green), or it may just slow down when the ego-car is going straight (orange). Based on the possible ego-car's future predictions, the other car's future object localization presents multiple modes. In Fig~\ref{fig:7b}, the white car can either turn to its left (green) or go straight (cyan) when the ego car is stopped. As highlighted in these examples, each modality of the uncertain ego-motion properly models interactive reactions of other agents using the proposed framework.

\section{Conclusion}
We introduced the NEMO framework to condition \textit{future object localization} on the uncertainty of future ego-motion priors. For this, we jointly modeled aleatoric and epistemic uncertainty of ego-motion prediction to sample multiple modes of future ego-behavior. Then, each modality was used as a prior to capture interactive reactions of other agents with respect to the different types of ego-motion. We also considered the uncertainty of future object localization as well as its multi-modality. To this end, ablative tests were conducted using the public benchmark dataset, comparing NEMO with the state-of-the-art methods and self-generated baseline models. We observed that combined epistemic and aleatoric uncertainty modeling in both \textit{future ego-motion prediction} and \textit{future object localization} achieved the lowest prediction error from both future ego-motion and future bounding box prediction. 
In the future, we plan to extend our work to rank each of the predicted modes based on the uncertainty measure, which will result in making the system more pragmatic for real world applications.

\bibliographystyle{IEEEtran}
\bibliography{reference.bib}
\end{document}